\pdfoutput=1

\documentclass[11pt]{article}

\usepackage[]{acl}
\usepackage{times}
\usepackage{latexsym}

\usepackage{algorithm}
\usepackage[noend]{algpseudocode}
\usepackage{graphicx}
\usepackage{multirow}
\usepackage{caption}
\usepackage{subcaption}
\usepackage{enumitem}
\usepackage{xspace}
\usepackage{subcaption}
\usepackage{comment}
\usepackage{booktabs}

\usepackage{amsmath}

\newcommand{\delphi}{Delphi\xspace}
\newcommand{\model}{\textsc{ClarifyDelphi}\xspace} 
\newcommand{\dataset}{\textsc{$\delta$-Clarify}\xspace} 

\newcommand{\socialchem}{\textsc{Social Chemistry}}
\newcommand{\moralstories}{\textsc{Moral Stories}}
\newcommand{\scruples}{\textsc{Scruples}}
\newcommand{\ethicsmoral}{\textsc{ETHICS} Commonsense Morality}

\newcommand{\sbic}{\textsc{Social Bias Inference Corpus}}
\newcommand{\unicorn}{\textsc{Unicorn}}

\newcommand\sit[1]{\textit{#1}}
\newcommand\ques[1]{``#1''}
\newcommand\ans[1]{``#1''}

\usepackage[T1]{fontenc}

\usepackage[utf8]{inputenc}

\usepackage{microtype}

\title{ClarifyDelphi: Reinforced Clarification Questions \\with Defeasibility Rewards
for Social and Moral Situations}

\newcommand{\aspace}{\hspace{1em}}
\newcommand{\uw}{$^{\heartsuit}$}
\newcommand{\aitwo}{$^{\spadesuit}$}
\newcommand{\biu}{$^{\diamondsuit}$}
\newcommand{\utah}{$^{\clubsuit}$}

\author{%
  Valentina Pyatkin\biu \aitwo \aspace
  Jena D. Hwang\aitwo \aspace
  Vivek Srikumar\aitwo \utah \aspace
  Ximing Lu\uw \aitwo \aspace \\
  \textbf{Liwei Jiang}\uw \aitwo \aspace
  \textbf{Yejin Choi}\uw \aitwo \aspace
  \textbf{Chandra Bhagavatula}\aitwo \aspace \\
    \biu{}Bar-Ilan University \:
    \utah{}University of Utah \\
      \aitwo{}Allen Institute for Artificial Intelligence \\
  \uw{}Paul G. Allen School of Computer Science \& Engineering, University of Washington \\
  \texttt{pyatkiv@biu.ac.il}
}

\begin{document}
\maketitle

\begin{abstract}
\sit{Context is everything}, even in commonsense moral reasoning. Changing contexts can flip the moral judgment of an action; \sit{Lying to a friend} is wrong in general, but may be morally acceptable if it is intended to protect their life.

We present \model, an interactive system that learns to ask clarification questions (e.g., \ques{why did you lie to your friend?}) in order to elicit additional salient contexts of a social or moral situation.   
We posit that questions whose potential answers lead to \textit{diverging} moral judgments are the most informative. Thus, we propose a reinforcement learning framework with a \emph{defeasibility reward} that aims to maximize the divergence between moral judgments of hypothetical answers to a question. Human evaluation demonstrates that our system generates more {relevant}, {informative} and {defeasible} questions compared to competitive baselines. 
Our work is ultimately inspired by studies in cognitive science that have investigated the flexibility in moral cognition (i.e., the diverse contexts in which moral rules can be bent), and we hope that research in this direction can assist both cognitive and computational investigations of moral judgments. 
\end{abstract}

\section{Introduction}

Commonsense moral reasoning of social situations and actions depends squarely on their context. \textit{Offering someone a cup of coffee} is generally considered appropriate. If offered to a work colleague, it may even be viewed as a courteous gesture. However, offering coffee to a toddler would be deemed morally irresponsible.

\begin{figure}[ht]
    \centering
    \includegraphics[width=0.45\textwidth]{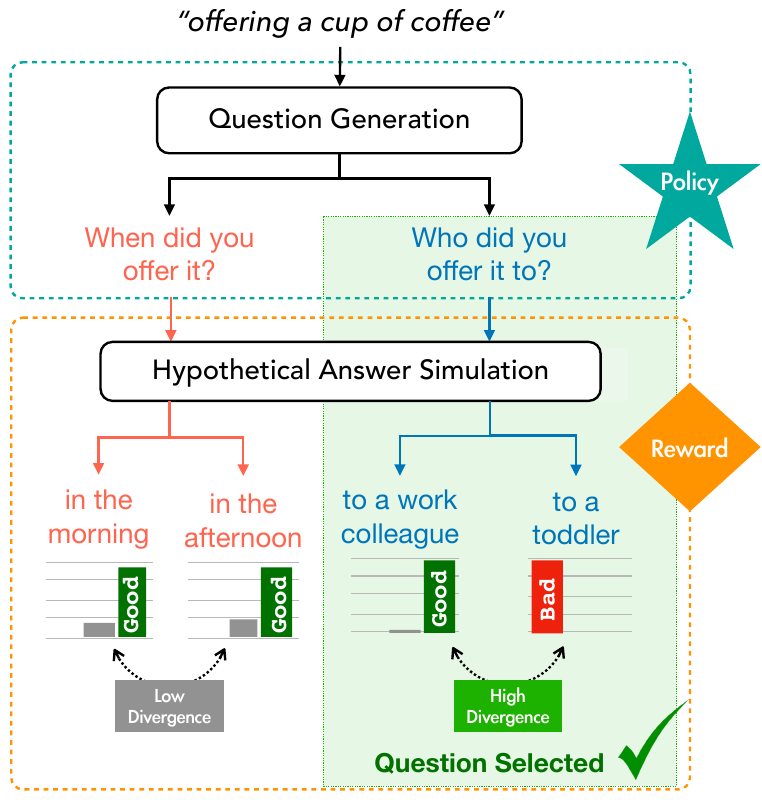}
    \caption{The \model{} question generation approach is trained via reinforcement learning. The reward simulates a set of possible (defeasible) answers to the questions and, using Delphi for feedback, optimizes for questions leading to maximally diverging answers.}
    \label{fig:first_fig}
\end{figure}

\begin{figure*}[ht!]
    \centering
\begin{subfigure}[T]{0.33\textwidth}
  \includegraphics[width=\linewidth]{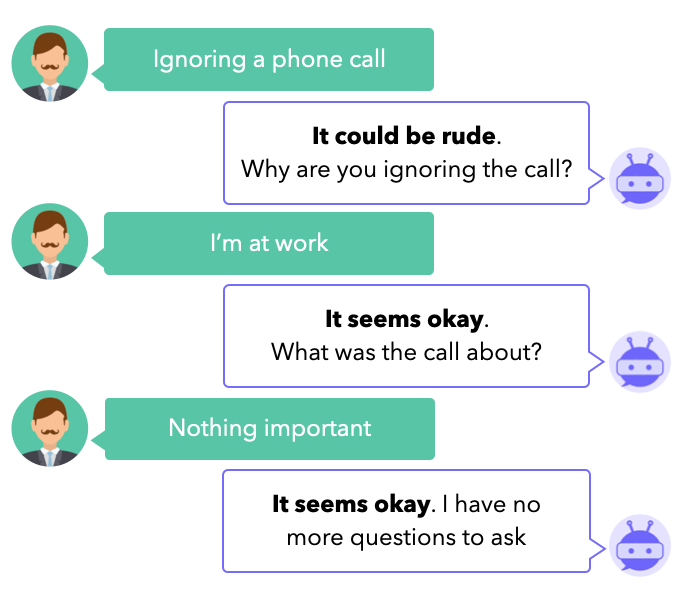}
  \caption{}
\end{subfigure}\hfil 
\begin{subfigure}[T]{0.33\textwidth}
  \includegraphics[width=\linewidth]{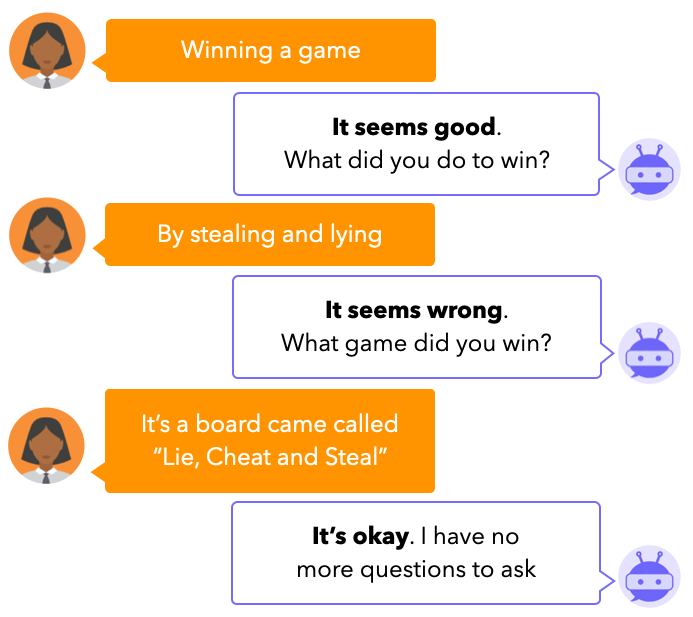}
  \caption{}
\end{subfigure}\hfil
\begin{subfigure}[T]{0.33\textwidth}
  \includegraphics[width=\linewidth]{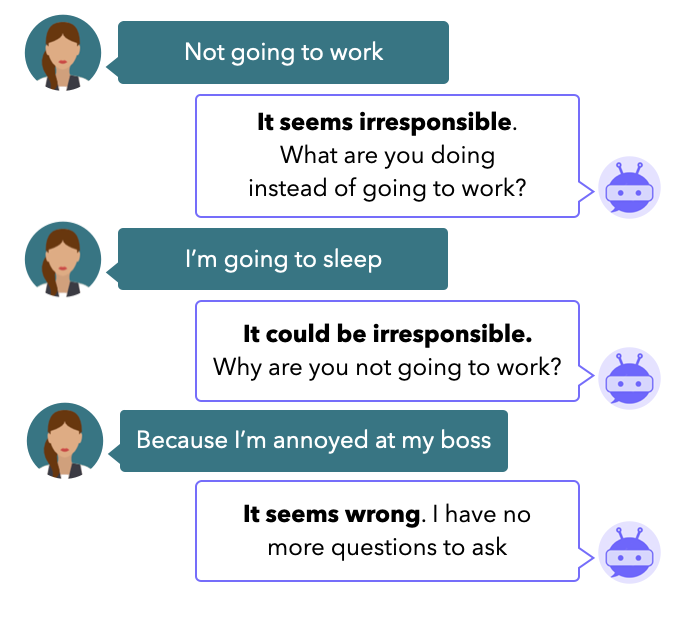}
  \caption{}
\end{subfigure}\hfil

    \caption{Interaction between a user and \model. The user inputs a situation and \model{} answers with an initial judgement (obtained from \textsc{Delphi}) and a clarification question, which the user then answers. }
    \label{fig:interaction}
\end{figure*}

Delphi \cite{jiang2022delphi}, a recently proposed commonsense moral reasoning model, generates moral judgments for simple actions described in text. However, Delphi's judgments are made in isolation, without any knowledge of surrounding context. Grounding moral reasoning in context is crucial \cite{talat-etal-2022-machine}. How can moral reasoners elicit missing salient context? A natural way to do so is by asking clarification questions.

We present \model, an interactive system that learns to ask questions to elicit salient context. Prior research in cognitive science shows that human reasoning exhibits the flexibility not only to articulate where a certain moral rule should hold, but also to imagine valid exceptions where the rule can be bent or \textit{defeated} based on the demands of the context \cite{kwon2022flexibility, levine2020logic, awad2022acceptable}. 

We present a first step toward computationally exploring
and discovering these \textit{defeasible} contexts  which can potentially flip the moral judgement of a situation. 
Given a situation and its default judgment (e.g., it is \textit{nice} to offer a cup of coffee to someone), defeasible contexts can strengthen (e.g., \sit{offering it to a colleague}) or weaken (e.g., \sit{giving it to a toddler}) the judgment
\cite{rudinger-etal-2020-thinking, madaan2021think, allaway2022penguins}.
We aim to generate questions whose answers might uncover missing context for making better-informed moral judgments, and we propose to 
do so in a conversational setting between a user and \model. 

Our method for clarification question generation is based on reinforcement learning. Using Proximal Policy Optimization (PPO; \citealt{schulman2017proximal, ouyang2022training}) we optimize for generating questions that 
invoke responses that provide morally salient contexts. 
\model ``imagines" answers to a generated question, using a trained answer generation model.  A reward is calculated by comparing the probability distributions Delphi assigns to the imagined answers. Fig.~\ref{fig:first_fig} provides an overview of \model.

The intuition behind our approach is that questions that lead to maximally divergent answers (e.g., \ques{Who did you offer it to?}) are also those that elicit most morally salient contexts and therefore are more \textit{consequential} to the situation. These morally consequential questions surface latent ambiguities that may directly affect the moral decision process.
Questions with little divergence in its imagined answers (e.g., \ques{When did you offer it?}) have little to offer in terms of resolving contextual moral ambiguities.

Our results show that our approach outperforms other strong clarification question generation baselines; its generated questions lead to 
consequential 
answers. We additionally quantify how much supervised clarification question training data is needed for a good initial policy. Lastly we show that questions help with generating defeasible updates.

Our contributions are as follows. We introduce the task of clarification question generation for social and moral situations. For this task we propose an RL based approach, defining defeasibility as a new type of relevance for clarification questions.
We publicly release \dataset, a dataset of 33k crowdsourced clarification questions, and \dataset$_{silver}$ containing generated questions conditioned on a defeasible inference dataset. We also release trained models with their code.\textbf{\footnote{Data and code are available at: \url{https://github.com/allenai/clarifydelphi}.}}
\section{Problem Setup}
\label{sec:problem-setup}
Given a situation, such as \sit{lie to my friend}, we aim to generate question(s) that are the most relevant 
for uncovering
the most consequential context with respect to making a social or moral judgement. While situations could evoke a multitude of potential questions, the following work is concerned with predicting questions whose answers are likely to be \textit{consequential}, i.e. answers that could function as either weakeners or strengtheners of the default judgement. 
The terms \textit{weakener} and \textit{strengthener} come from the concept of defeasible inference \cite{rudinger-etal-2020-thinking}, which defines a way of reasoning that takes into consideration (new) evidence which could either support (e.g. \textit{strengthen}) or cancel/\textit{weaken} an initial inference.

Formally, the task is to predict a question $q$ given a base situation $s$. The base situation has a default moral judgement $j\in\{bad, ok, good\}$. For every input tuple of $(s_i, q_i, j_i)$ there is a hypothetical set of strengthening answers $A_S$ and weakening answers $A_W$. Adding the additional information obtained from any $q_i$ and corresponding answer $a_i$ to the base situation $s_i$ results in an updated situation $s_{ui}$, with an updated judgement $j_{ui}$.

\section{\model: A Reinforced Clarification Question Generator}

\begin{figure}[t]
\begin{algorithm}[H]
\footnotesize
\caption{
    Training \model
}
\textbf{Input} initial policy model $\theta_0$, initial value model $\phi_0$, \delphi $\psi_{\text{Delphi}}$\\
 \hspace*{2mm} $D_{\dataset} \gets$ Get dataset of clarification questions. \\
  \hspace*{2mm} $\theta_{\text{Q}} \gets$ Fine-tune $\theta_0$ with Eqn~\ref{eqn:qgen_policy} from $D_{\text{\dataset}}$. \algorithmiccomment{Sec.~\ref{sec:policy}} \\
   \hspace*{2mm} $D_{\dataset_{silver}} \gets$ Get silver dataset of defeasible answers to questions. \\
     \hspace*{2mm} $\theta_{\text{A}} \gets$ Fine-tune lm with Eqn~\ref{eqn:agen_policy} from $D_{\text{\dataset}}$. \algorithmiccomment{Sec.~\ref{sec:answersim}} \\
   \hspace*{2mm} $\theta_{\model{}} \gets$ \Call{ReinforcedLearning}{$S_{\text{SocialChem}}$, $\theta_{\text{Q}}$, $\theta_{\text{A}}$, $\phi_0$, $\psi_{\text{\delphi}}$} \\
\hspace*{2mm} \phantom{...} \algorithmiccomment{Sec.~\ref{sec:rl}} 
\begin{algorithmic}
\Procedure{ReinforcedLearning}{$S_{\text{SocialChem}}$, $\theta_{\text{Q}}$, $\theta_{\text{A}}$, $\phi$, $\psi_{\text{\delphi}}$} 
    \State $\theta_{Q_{\text{old}}} \gets \theta_Q$, $\phi_{\text{old}} \gets \phi$
    \For{iterations = 1, 2, \ldots}
        \State Sample a minibatch of situations $s$ from $S_{SocialChem}$.
        \For{step = 1, 2, \ldots, $s$}
            \State Calculate $r$ using $\theta_A$ and $\psi_{\text{Delphi}}$ with Eqn~\ref{eqn:reward}.
            \State Compute $loss_{\text{PPO}}$ on the minibatch with Eqn~\ref{eqn:loss}.
            \State Optimize $\theta$ and $\phi$ with $\mathcal{L}_{\text{PPO}}$ for one step.
        \EndFor
        \State $\theta_{Q_{\text{old}}} \gets \theta_Q$, $\phi_{\text{old}} \gets \phi$
    \EndFor
    \State \Return $\theta_Q$
\EndProcedure
\end{algorithmic}
\textbf{Output} $\theta_{\model}$
\label{alg:ppo}
\end{algorithm}
\end{figure}

The \model approach is based on reinforcement learning. Algorithm \ref{alg:ppo} gives an overview of the training process.
As a first step, before performing reinforcement learning, we obtain a question generation model $\theta_Q$ and an answer generation model $\theta_A$, which we both train on data that we curated, described in the later Sec.~\ref{sec:dataset}. The question generation model predicts the clarification questions and the answer generation model provides (defeasible) answers to the generated questions.
By using these two models in addition to \delphi ($\psi_{\delphi}$) for calculating the rewards, we do not require any supervised data during RL training. 

We consider question generation conditioned on a given situation a sequential decision making process over the natural language vocabulary space, where the generated question $q$ with $T$ tokens has an episode of length $T$. At step $t \in [1, T]$, the state $s_t = (s, q_{<t})$ is the combination of the given situation and the question decoded up to the $(t-1)$-th token; the action $c_t = q_t$ would be the $t$-th token to decode.
The question generation model, $\theta_Q(q_t | q, q_{<t}; \theta)$, is
the \textit{policy model} that we optimize.
We define a reward function $r(s, q, a_w, a_s)$ that characterizes the divergence of answers from $\theta_A$ conditioned on generated question $q$ and discuss the definition of this reward function in \S\ref{sec:reward}.

\subsection{Supervised Question Generation}
\label{sec:policy}
 The first subcomponent is a basic question generation system $\theta_Q$ that outputs a question $q$ conditioned on a situation $s$. It is used as the initial policy model during RL training.
 \begin{align}
    \hat{q}=\underset{q}{arg\,max} P(q|s)
\label{eqn:qgen_policy}
 \end{align}

\subsection{Defeasible Answer Simulation}
\label{sec:answersim}
For each generated question $q$, we need to generate a weakening answer $a_w$ and a strengthening answer $a_s$ in order to calculate the reward $r$ (Formula~\ref{eqn:reward}).
For the defeasible answer generation system $\theta_A$, we take as input a situation $s_i$, the generated question $q_i$ (\S\ref{sec:policy}), and an update type $u \in \{weakener, strengthener\}$ to predict a weakener-strengthener answer pair $a_w$ and $a_s$:
\begin{align}
a = \underset{a}{arg\,max} P(a|s, q, u)\label{eqn:agen_policy}
 \end{align}

An example of an instantiated input/output:
\begin{description}
  \setlength\itemsep{-0.4em}
   \item[Input] It's bad to be a snitch, TYPE: Weakener, Q.: Why would being a snitch be beneficial?
   \item[Output] doing so would save someones life.
\end{description}

The crucial element in the input is the update type, as it allows to generate two types of answers for the same $s$ and $q$.
When computing the reward during training, for each question, we filter out all its generated answers which either contradict or are entailed (i.e. no new information) by the given situation, using an off-the-shelf NLI model.

\subsection{Reward}
\label{sec:reward}
As a reward for generating a question, we aim to quantify how well the generated questions are able to elicit consequential answers. For this purpose we query Delphi \cite{jiang2022delphi} for feedback, using situations updated with answers.

We optimize for questions that lead to maximally divergent answers by defining a reward function which uses the JS-Divergence, between the Delphi probability distribution of the weakener updated situation and the strengthener updated situation:

\vspace{-1em}
 \begin{align}
    r(s, q, a_w, a_s) = JSD(P_{jw}||P_{js})
\label{eqn:reward}
 \end{align}

\paragraph{Sentence Fusion} 
To create an \textit{updated situation} that sounds natural and can be used to query Delphi, the situation $s$, question $q_i$ and answer (both $a_w$ and $a_s$ separately) have to be fused together into $s_{ui}$. 
For example:\\
\textbf{Situation} refraining from doing something bad\\
\textbf{Question} When do you do something bad?\\
\textbf{Answer} when I'm angry\\
\textbf{Fusion:} \sit{refraining from doing something bad when you're angry}.\\ 
We train a model to distill fusion in-context examples obtained from GPT-3 (text-curie-001).

\paragraph{Delphi for Feedback}
Delphi is then queried with the updated situation $s_{ui}$ for a judgement,
leveraging the probability distribution that Delphi provides
over three classes: $j\in\{\text{bad}, \text{ok}, \text{good}\}$. 
The probability scores are the probabilities of the special T5 tokens representing each of the three classes, normalized by their sum.
 \begin{align}
    j = \underset{q}{arg\,max_j} P(j|s)
\label{eqn:delphi_probs}
 \end{align}

\paragraph{JS-Divergence}
We calculate the Jensen-Shannon divergence between the Delphi probability distributions $j_w$ and $j_s$ obtained from two updated situations originating from defeasible answers to $q$.

\paragraph{Reward normalization}
We normalize the reward during training as follows: 
\begin{align}
r(x, k) &\leftarrow \frac{r(x, k) - \mu_0}{\sigma_0}.
\label{eqn:reward-norm-apply}
\end{align}

The $\mu_0$ and $\sigma_0$ of rewards are obtained before training begins, by generating a question and calculating its rewards for all $s$ in the training data.

\subsection{Proximal Policy Optimization (PPO) }
\label{sec:rl}

We maximize the reward using Proximal Policy Optimization (PPO) \cite{schulman2017proximal} as our RL algorithm, which previous works have shown to be suitable for NLG tasks \cite{liu2022rainier, ramamurthy2022reinforcement}. Our implementation of PPO is an adaptions of \citet{ouyang2022training}'s, which includes a KL penalty between the initial policy model $\theta_{Q_{old}}$ and the updated policy $\theta_Q$. 
In addition to the policy model, PPO employs a \textit{value model} (parameterized by $\phi$) to estimate the value function for states with incomplete decoded text, i.e. $V(s_t; \phi)$ for any $t$.
PPO's loss consists of a value model ($\phi$) loss and the policy loss, which is jointly minimized: 
\begin{align}
loss_{\text{PPO}}(\theta, \phi) &= \alpha \cdot loss_{\text{Value}}(\phi) + loss_{\text{Policy}}(\theta)
\label{eqn:loss}
\end{align}

\section{\dataset: a Dataset of Clarification Question}
\label{sec:dataset}

\begin{figure}[t]
    \centering
    \includegraphics[width=0.5\textwidth]{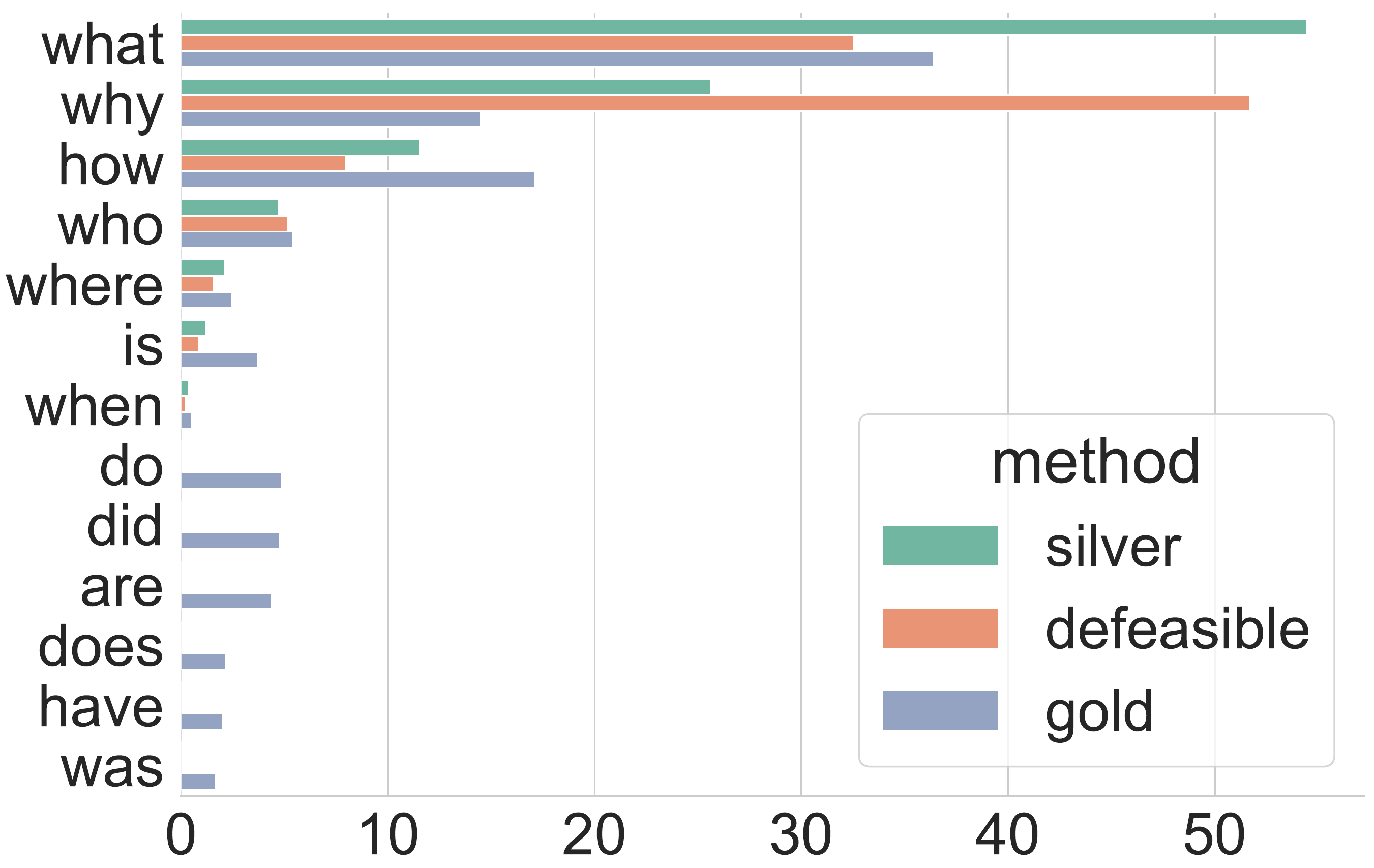}
    \caption{Proportional distribution (\%) of the most frequent question starts in $\dataset_{gold}$, $\dataset_{silver}$ and the subset of \textit{defeasible} questions of $\dataset_{silver}$.}
    \label{fig:qstarts}
\end{figure}

\begin{table}[b]
\centering
\small
\begin{tabular}{p{.7cm}p{6cm}}
\toprule
\multicolumn{2}{l}{\textit{Situation:} \textbf{Tipping people decently}}  \\
 Q1 & What did they do for you?     \\   
 Q2 & Can you afford to tip?    \\ 
 Q3 & Was the service good?  \\
 Q4 & Did the people perform the service adequately?     \\              
 Q5 & Do you always tip people well regardless of the service quality? \\
 
\midrule

\multicolumn{2}{l}{\textit{Situation:} \textbf{Jeff ignores the comment and }} \\                
\multicolumn{2}{l}{\hspace{14mm}\textbf{laughs about it with his boss}} \\                
   Q1-4 & What was the comment?     \\                                          
   Q5 & Who made the comment they were laughing at? \\
   
   \bottomrule
   
\end{tabular}
    \caption{Two examples of situations and their clarification questions, written by five different Turkers.}
    \label{tab:datasetexamples}
\end{table}

We require data for various components of our \model model: The policy needs bootstrapping from a clarification question dataset and the answer generation model needs data to learn to generate defeasible answers to questions. To the best of our knowledge no such datasets exist.
We therefore collect a crowdsourced dataset of clarification question for social and moral situation and a silver dataset of defeasible QAs to train $\theta_Q$ .

The situations are sampled from \textsc{Social-Chem-101} \cite{forbes2020socialchemistry} and the \textsc{Commonsense Norm Bank} \cite{jiang2022delphi}. We call our dataset \dataset and it consists of crowdsourced questions, enriched with questions generated by GPT-3 \cite{brown2020language}.

Next we describe how we create the dataset.

\paragraph{$\dataset_{gold}$:}
We crowdsource clarification questions by
showing annotators a situation and asking them to write a clarification question they would ask an imagined colleague requesting advice on the situation.
Each of the 6425 situations is presented to 5 annotators, resulting in 5 questions per situation (500 situations are used for dev and test respectively). Details of the annotation are found in Appendix~\ref{sec:annotation}.

\paragraph{$\dataset_{silver}$:}
\label{silver}
The $\delta$-SOCIAL part of the defeasible inference dataset \cite{rudinger-etal-2020-thinking} consists of 
\textit{statements} that express default judgments over situations (\sit{It is good to protect your kids}) 
and \textit{updates} 
that weaken (\sit{Your protection is overbearing}) or strengthen (\sit{Your child is in danger}) the default. 
These updates could be viewed as potential answers to an implicit question about a base situation: \ques{What are you protecting your child from?} 
for \sit{Your child is in danger.} We 5-shot prompt GPT-3 to generate questions, conditioned on situation and answer, resulting in  $\approx80K$ $(situation, update \,type, question, answer)$ tuples.

\paragraph{Dataset Analysis}
Fig.~\ref{fig:qstarts} shows that the crowdsourced $\dataset_{gold}$ has more variety in its most common question starts, which reflects the general diversity of the dataset: For only 10\% of the situations, more than 1 Turker asked exactly the same question, and for only 8\% of the situations all 5 Turkers used the same Wh-word to start the question. This 
highlights that
there is more than one possible salient clarification question to be asked for any given situation. For the situation \sit{tipping people decently} in Tab.~\ref{tab:datasetexamples}, all 5 Turkers chose to start their questions differently, even though three out of these five questions ask in one way or the other about the service quality. For the other situation 4/5 Turkers asked for a specification \ques{What was the comment?} and 1 Turker asked about the missing agent role. We also see that polar (yes/no) questions appear less frequently, 
as Turkers were explicitly asked to avoid them
unless no other suitable question comes to mind.\footnote{This is to prevent leading questions such as ``Do you intend to give it to a kid?" for "offering a cup of coffee".
}

The $\dataset_{silver}$ questions are generated by conditioning on weakener or strengthener updates. Since we aim to predict \textit{defeasible} questions, the most desirable questions are those whose answers can be both weakeners and strengtheners. In the silver data, 53\% of situations have at least one question that has been generated by GPT-3 for both update types. The situation \sit{Your kids should be your number one priority}, for example, has the same question \ques{What are your kids' ages?} for the weakener update \sit{They are adult children.} and the strengthener update \sit{Your children are toddlers.}
Interestingly, among the 
subset of \textit{defeasible} questions in $\dataset_{silver}$, we find that the most frequent question start is `why'. This suggests that it is easiest to come up with both weakener and strengthener answers to why-questions. 
\section{Baselines}
We consider four baselines in our experiments.

\paragraph{Question Generation Without RL}
To assess what additional improvements training an RL model with a defeasibility rewards provides, we report performance of the supervised question generation model $\theta_Q$ on its own (\S\ref{sec:policy}). We refer to this baseline as \textit{t5 fine-tuned}. We decode using nucleus sampling with top-\textit{p} $=0.6$.

\paragraph{Pipelines with Question Selection}
Next, we implement two pipelines where, as the first step, a diverse set of questions is generated for a given situation and then, as the second step, the best question is selected according to a score.

In order to generate a diverse set of questions we fine-tune T5 on \dataset, conditioned on a modified input compared to the model from \S\ref{sec:policy}: \textbf{Input} <Situation>. Q.: <wh-word> - \textbf{Output} <Question>

By also conditioning on the first wh-word of the question it is possible to generate different questions. During inference we generate questions for 14 different question starts.\footnote{\textit{what, how who, do, are, did, is where, have, was when, would}}
We propose two approaches to scoring these questions: using a discriminator model and using divergence ranking, which we describe as follows.

\paragraph{Discriminator}
We train a discriminator classifier which labels these questions as either \textit{relevant} or \textit{irrelevant} to a given situation. We then choose the question that has been assigned the \textit{relevant} label with the highest probability.

The discriminator is a binary classifier based on DeBERTa \cite{he2020deberta}. The positive examples are situations and their respective 5 questions written by annotators. The negative question examples are sampled from other situations, in a way that ensures that the \textit{relevant} and \textit{irrelevant} questions are similar enough to make training more challenging.

\paragraph{Divergence Ranking}
We run the defeasible answer simulation with feedback from Delphi for each question in the set. This process is the same as the reward function of the RL approach, except that the JS-divergence score is used to rank the questions instead of being used as a reward for question generation. We compare two variations of this baseline: one with answer filtering using an NLI model as described in Sec.~\ref{sec:answersim} (\textit{pipeline-nli}) and one without filtering (\textit{pipeline}).

\paragraph{Why-Baseline}
We saw in \S\ref{sec:dataset} that questions conditioned on weakener/strengthener updates are usually causal questions.
Using the same input/output configuration as in the pipeline baseline, we generate a why-question for each situation (called \textit{why}).
\section{Evaluation and Analysis}
\subsection{Human Evaluation}
\label{sec:human_eval}
Automatic evaluation of questions and their usefulness for clarifying moral situations is tricky. While we do have gold reference questions, we have shown that humans will produce diverse questions for the same situation (\S\ref{sec:dataset}) and just because a question does not appear in the reference set does not necessarily indicate that it is not a consequential question.
We therefore perform human evaluation of the models' outputs on Amazon Mechanical Turk on the 500 test set instances from \dataset.
Given a situation and a question, Turkers are asked to rate the question along three different attributes:  \textbf{Grammaticality} (Is the question grammatical?), \textbf{Relevance} (Is the question relevant and plausible to the situation?), and \textbf{Informativeness} (Does the question access new information or regurgitate the situation?). The attributes are further detailed in Appendix~\ref{sec:annotation}.

Additionally, and most importantly, we aim to evaluate the \textbf{defeasibility} of the questions, e.g. how well the generated questions can elicit \textit{weakener} or \textit{strengthener} answers.
For this purpose, Turkers are given a situation with a question and are first 
asked to judge this situation (\textit{generally ok} vs. \textit{generally not ok}). They then need to say whether 
and specify if they can think of an answer to the question which might \textit{support} their judgement and 
also of
an answer which would \textit{flip} their judgement.

\subsection{Results of Human Evaluation}
We first run the \textit{grammaticality}, \textit{relevance} and \textit{informativeness} evaluation. All questions which are given the lowest rating (e.g. \textit{irrelevant} and/or \textit{uninformative}) by at least two annotators are excluded from the second evaluation. 
It does not make sense to ask about defeasibility for questions which already are \textit{irrelevant}, and additional weakening or strengthening context is not feasible for \textit{uninformative} questions.

\begin{figure}[]

    \centering
    \includegraphics[width=0.5\textwidth]{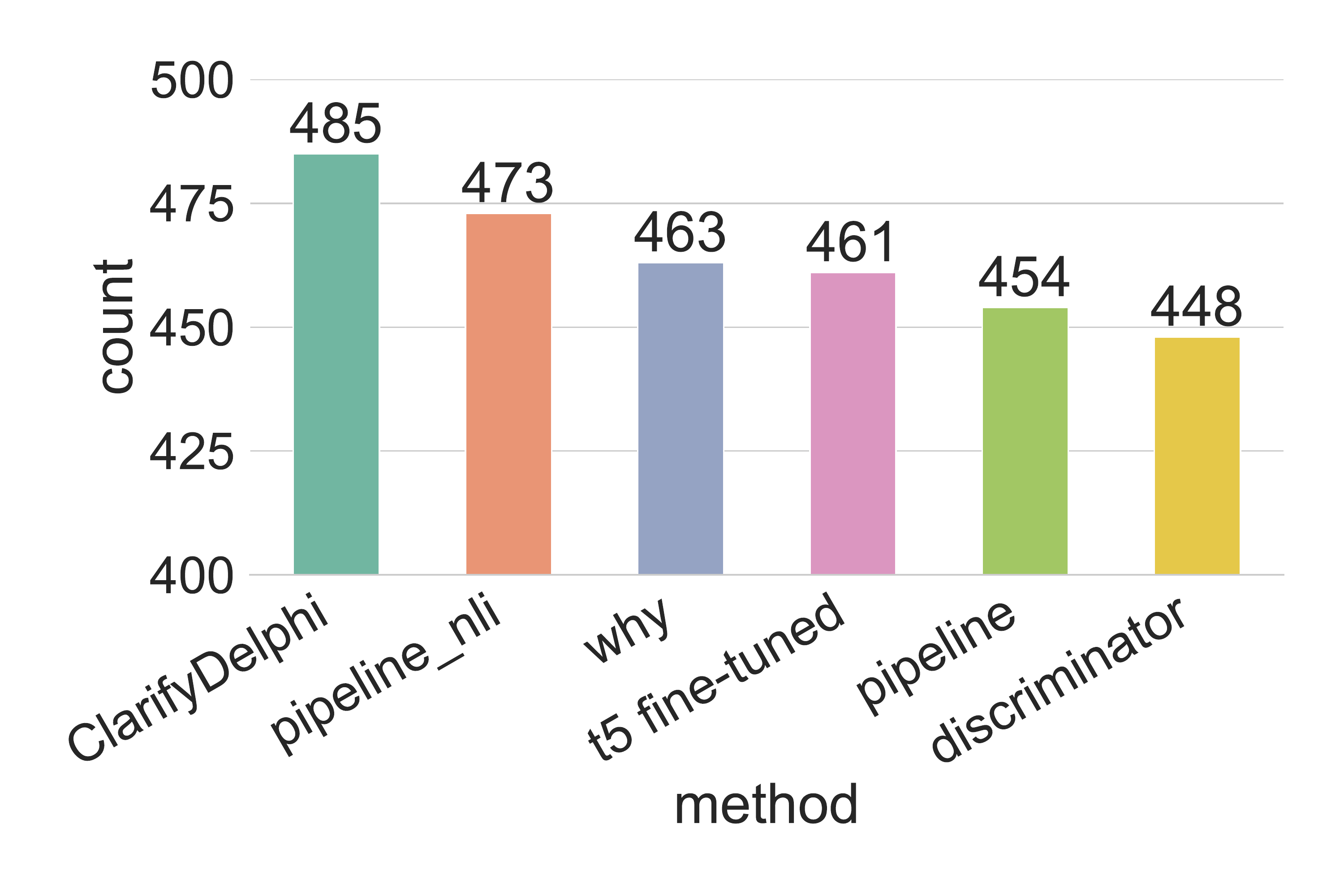}
    \caption{Number of questions (out of 500) in test set that received an 
    \textit{informativeness} and \textit{relevance} rating of $>0$.}
    \label{fig:releval}
\end{figure}

We find, as displayed in Fig.~\ref{fig:releval}, that \model has the biggest percentage of \textit{relevant} and \textit{informative} questions in the test set, compared to the baselines. We also see that a big majority of the generated questions, from all models, are \textit{relevant} and \textit{informative}, with the lowest performing model (\textit{discriminator}) still producing 448/500 questions that are passed on to the next evaluation round.

We also find that \textit{grammaticality} across all systems is high with the lowest average score being 0.98 and the highest 0.99 (on a scale from 0 to 1, with 1 being grammatical). The minimal variation in grammaticality score is expected since all models are based upon the same transformer model.

The \model questions also outperform the baselines in terms of defeasibility, as seen in Table~\ref{tab:defeasibility}: annotators can more often think of a \textit{strengthener} answer and/or a \textit{weakener} answer to our questions. 
The evaluation also shows that adding the answer-filtering with NLI step to the pipeline improves the question selection on all 4 evaluation criteria. The why-baseline is shown to be a strong baseline, indicating that motives and reasons are important for moral reasoning.

\begin{table}[ht]
\small
\begin{tabular}{l|lll}
              & defeasibility & weakener & strength. \\ \hline
\model      & \textbf{0.44}          & \textbf{0.47}     & \textbf{0.73}         \\
why           & 0.37          & 0.41     & 0.60         \\
pipeline\_nli & 0.35          & 0.37     & 0.64         \\
t5 fine-tuned & 0.34          & 0.37     & 0.54         \\
discriminator & 0.33          & 0.36     & 0.55         \\
pipeline      & 0.30          & 0.34     & 0.53        
\end{tabular}
    \caption{Defeasibility scores obtained through human evaluation. \textit{weakener}: Can you think of an answer to the question that weakens your initial judgement? \textit{strengthener}: Can you think of an answer ot the question that strengthens your intial judgement? \textit{defeasibility}: Can you think of both?.}
    \label{tab:defeasibility}
\end{table}

\subsection{How much supervision does the policy require?}
\label{sec:automatic_eval}
Our approach uses RL in conjunction with a supervised policy that has been fine-tuned on question generation. This has been shown to outperform approaches which use RL on top of a ``vanilla" lm-policy \cite{ramamurthy2022reinforcement}. To assess the effect of supervision on question generation performance, we trained multiple initial policies on varying percentages of \dataset training data: 25\%, 50\%, 75\% and 100\%. To compare against more traditional supervised question generation approaches we additionally trained a policy on SQuAD v1.1 data \cite{rajpurkar2016squad}.

We report two automatic evaluation metrics. To measure \textit{informativeness} we use an off-the-shelf QA model trained on SQuAD 2.0 from AllenNLP \cite{gardner2018allennlp}. This model either answers a question by pointing to a span in the input or outputs that the question in unanswerable with respect to a given context. For a clarification question to be informative it would not ask about anything already mentioned in the situation. For the \textit{QA}-metric we thus report the percentage of non-answerable questions.\footnote{The Pearson correlation coefficient shows that this metric (moderately) correlates with the human informativeness evaluation ($r=0.31$).}
We also report the average maximum BERTScore \cite{zhang2019bertscore} between a generated question and one of the 5 gold questions in \dataset.

\begin{figure}[ht]
    \centering
  \includegraphics[width=\linewidth]{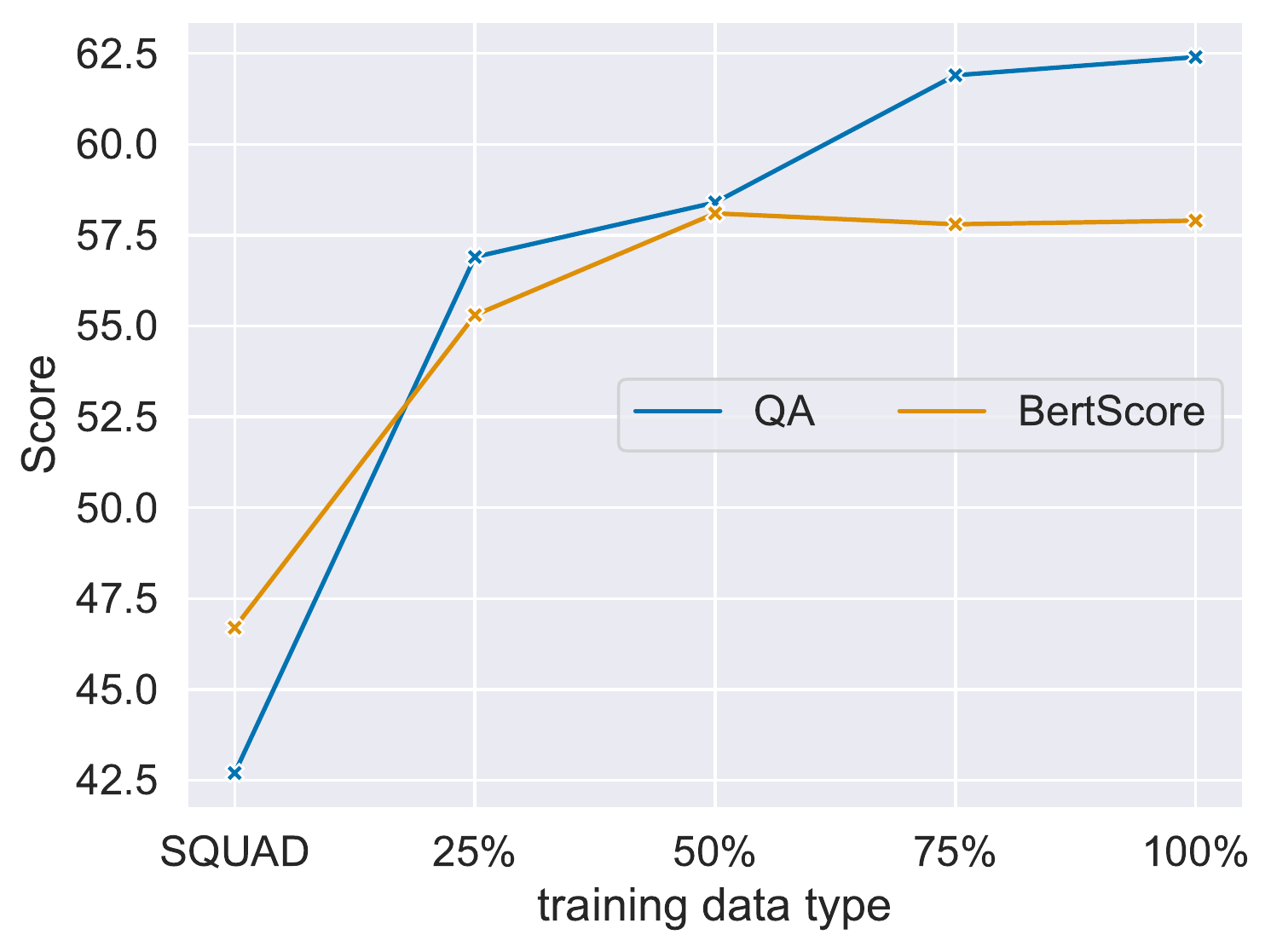}
    \caption{Performance of ppo algorithm with different policies: a policy pre-trained on SQUAD and policies pre-trained on different subsets of the \dataset dataset. The scores (higher is better) are averaged every 1000 steps, between 1000 and 6000.}
    \label{fig:training_data_curves}
\end{figure}

Fig.~\ref{fig:training_data_curves} shows the following trends with regards to training a supervised policy. More training data leads to more informative questions. The policy trained on SQuAD produces the most uninformative questions which can be explained by the fact that SQuAD questions are conditioned on existing answers in a text. While performance consistently increases from 25\% to 75\% of the training data, improvements after 75\% are minimal. We conclude that for our use case training on about 5000 (75\%) situations with 5 questions each leads to a sufficiently good policy. These results are also supported by the BERTScore.

\subsection{Analysis}

\paragraph{Answer Simulation}
The answer generation model generally succeeds at generating diverse \textit{weakener} and \textit{strengthener} answers to the same question: for only about 0.05\% of questions per 1000 PPO epochs the model generates the same answer for both weakener and strengthener. 

Our answer generation could be looked at as question-guided defeasible update generation. \citet{rudinger-etal-2020-thinking}'s task of \textit{Generative Defeasible Inference} generates an update given a situation, a moral judgement and the update type (e.g. weakener/strengthener).
In our answer generation approach we condition on the same input together with a generated question. This intermediate question generation step functions as a type of macro planning which has been shown to be effective for NLG \cite{puduppully-etal-2022-data, narayan2022conditional}. We evaluate our approach on the same test set using the same evaluation metrics as \citet{rudinger-etal-2020-thinking}. Table~\ref{tab:answer_gen_eval} shows that by first predicting the question and then the updates, we improve upon generating defeasible updates for $\delta$-\textsc{social}.

\begin{table}[t]
\small
\centering
\begin{tabular}{l|ll}
                               & BLEU  & ROUGE \\ \hline
$\delta$-\textsc{social} (T5-large)        & 4.22  & 14.94 \\
$\delta$-\textsc{social} (GPT2-XL)         & 12.16 & 18.77 \\
\textsc{ClarifyD} (T5-large) & \textbf{14.18} & \textbf{34.65}
\end{tabular}
\caption{Automatic results for strengthener/weakener update generation on the $\delta$-\textsc{social} test set. Following \citet{rudinger-etal-2020-thinking} we report BLEU-4 \cite{papineni2002bleu} and ROUGE-L \cite{lin2002manual}. The first two results are from \citet{rudinger-etal-2020-thinking}.}
\label{tab:answer_gen_eval}
\end{table}

\paragraph{Questions}

We qualitatively inspect the types of generated questions:
There are many specification questions asking about a hyponym of an argument in the base situation, for example, \sit{exterminating pests on your property} - \ques{What kind of pests?}. The situations extracted from SocialChem often include underspecified pronouns, such as `something' or `somewhere'. 60\% of the situations containing `something', for example, elicit \textit{what}-questions from our model. Note that while such specification questions are valid clarification questions, the SQUAD 2.0 QA model would mark them as answerable given the situation. 
It is also interesting to see that often when a situation has a missing or implicit semantic argument, such as \sit{being anxious sometimes}, \model inquires about it: \ques{What are you anxious about?}
The generated \textit{why}-questions most often ask about the motives of the agent in the situation, such as \sit{Ben tells Maggie that he's traveling alone} - \ques{Why is Ben traveling alone?}. More rarely the model generates questions asking about the viewpoint of the patient: \sit{asking a friend [...] whether your attire is appropriate for an event} - \ques{What is the friend's opinion of your attire?}

\paragraph{Analysis of Delphi's Probabilities}

In Tab.~\ref{tab:defeasibility_automatic} we quantify the JSD of Delphi's judgments. Even though the human evaluation showed that \model produced the most questions leading to defeasible answers, the JSD and the precentage of judgment flips is higher for the pipeline\_nli approach, where we explicitly filter questions to maximize the JSD. Nevertheless, \model leads to more Delphi judgment flips and higher JSD between answers than the fine-tuned t5 model without RL (and also all other baselines besides the pipeline). This automatic evaluation and the disagreement with the human annotators also reveals that Delphi's probabilities are not always perfectly calibrated and relying too much on a model's output might potentially lead to some error propagation.

\begin{table}[ht]
\small
\begin{tabular}{l|lll}
              & avg. JSD & Judgment Flips \\ \hline
\model      & 0.191          & 25\%   \\
why           & 0.159          &   22\%       \\
pipeline\_nli & 0.259          & 33\%          \\
t5 fine-tuned & 0.144          & 21\%         \\
discriminator & 0.138          & 21\%          \\
\end{tabular}
    \caption{Average JSD between $P_{jw}$ and $P_{js}$ of a situation. Judgment Flips: \% of answers which led to a flip in Delphi's judgment.}
    \label{tab:defeasibility_automatic}
\end{table}
\section{Interactive Judgements}
\label{sec:interactive-judgments}

While we use answer simulation during PPO training, inference only requires a situation as input. The clarification questions can then be used to elicit additional context, in the form of answers, through interaction. Fig.~\ref{fig:interaction} illustrates examples of such an interaction between a user, Delphi as the moral reasoning system and \model. 
After each turn, the situation is updated with the user provided context, for which Delphi produces a new decision. 
We limit the interaction to three turns. This is based on the observation that after the third turn the sentence fusion starts to deteriorate, resulting in less relevant and more repetitive questions. 
Additionally, we find that the first two questions 
generally
can capture missing contexts that are most central to making moral decisions. We provide more examples of generated questions in the Appendix.
\section{Related Work}
\paragraph{Question Generation}

Clarification question generation has been studied for various domains from image recognition questions to product description questions \cite{rao2018learning, majumder2021ask, white2021open}, defining the goodness of clarification questions along the lines of information theoretic measures such as relevance, informativeness or utility \cite{rao2018learning, white2021open, warstadt2022testing, rao2018learning, rao2019answer}. Most of existing works focus on questions that lead to single true answer, whereas we focus on generating clarification questions based on social situations, defining the relevance and utility of a question in terms of defeasibility. Additionally, we offer a high-quality clarification question dataset for social and moral situation---comprising of more than 30K questions---that breaks the mold from the domain-specificity of previous clarification datasets \cite{kumar2020clarq, aliannejadi2019asking}.

Some general question generation approaches have incorporated an RL-based approach. \citet{buck2018ask} learn to paraphrase questions with a reward that maximizes the QA answer F1 score. And \citet{rao2019answer} optimize a binary utility reward, using \textit{Reinforce} in an adversarial setup for generating clarification questions. In our setup, we use Proximal Policy Optimization \cite{schulman2017proximal, ouyang2022training} with a trained model for feedback as part of the reward.

\paragraph{Commonsense Moral Reasoning}
Delphi \cite{jiang2022delphi} is a commonsense moral reasoning model trained on \textsc{Commonsense Norm Bank}, a dataset with 1.7M instances of descriptive knowledge of people's general sense of what's ethically acceptable or not in everyday situations. \textsc{Commonsense Norm Bank} is compiled from five existing large-scale datasets, including \socialchem{} \citep{forbes2020socialchemistry}, \ethicsmoral{} \citep{hendrycks2021aligning}, \moralstories{} \citep{emelin2020moral}, \sbic{} \citep{sap2020socialbiasframes}, and \scruples{} \citep{lourie2021scruples}.

Delphi is based on pre-trained \unicorn{}, a universal commonsense reasoning model, trained on a number of commonsense reasoning tasks.
Delphi can predict the ethical judgment given a description of a situation.

\section{Conclusion}
In this work we introduce \model, which generates clarification questions for social and moral situations. We show how a RL approach that optimizes for maximally divergent answers in terms of defeasibility outperforms other clarification question baselines. While we start with a supervised policy, the reward function makes use of already trained models and does not rely on any additional training data.
We believe that our questions can be useful for providing more disambiguating context through interaction.

\section*{Limitations}\label{sec:limitations}
\paragraph{On Western-centricity} The majority of the crowdworkers producing the source data ($\delta$-Social and Delphi) and \dataset were located in the United States. Due to this, the predictions generated by \model are currently limited to representing only the perspectives of western culture (particularly the United States). Overcoming the western-centric bias is a compelling direction for future research. 

\paragraph{On Defeasibility} We rely upon Delphi to produce acceptable judgments given a situation and the modifying context as a measure of defeasibility. We recognize that, however, Delphi is not perfect and is characterized by a variety of limitations such as limited cultural awareness and inconsistent predictions \cite{jiang2022delphi}. Investigating improved methods for identifying answer divergences that will better capture defeasibility is a topic for future investigation.

\section*{Ethics Statement}\label{sec:ethics}

Annotations are conducted on Amazon Mechanical Turk (MTurk). We maintain an average pay of \$15 per hour for all our crowdsourcing data collection and evaluation tasks. Our crowdsourcing tasks do not collect personal information and are strictly limited to gathering workers' general knowledge. We do not keep any deanonymizing information such as MTurk IDs so that the identity of the workers cannot be directly or indirectly ascertained. Finally, our crowdsourcing task meets the standards for exemptions as human research and has obtained the necessary documentation that deems it exempt through our internal IRB procedures.

Our model is intended to be used for research purposes only and it is not supposed to provide any sort of advice applicable outside of that domain.

\section*{Acknowledgements}
We thank our colleagues on the Beaker Team at the Allen Institute for AI for helping with the compute infrastructure. This work was supported in-part by DARPA MCS program through NIWC Pacific (N66001-19-2-4031).

\bibliography{anthology,custom}
\bibliographystyle{acl_natbib}

\appendix

\section{Appendix}
\label{sec:appendix}

\subsection{Crowdsourcing and Annotation}
\label{sec:annotation}

Annotations are collected on Amazon Mechanical Turk (MTurk). We run two varieties of HITs: (1) data collection HIT in which we collect questions given statements, and (2) evaluation HIT in which the workers are asked to judge validity of the generated questions. The group of 145 Turkers working on the HIts were manually vetted and selected through an open paid qualification round. We maintain an average pay rate of \$15/hour for all HITs.

\begin{figure*}[h!]
    \centering
    \includegraphics[width=0.9\textwidth]{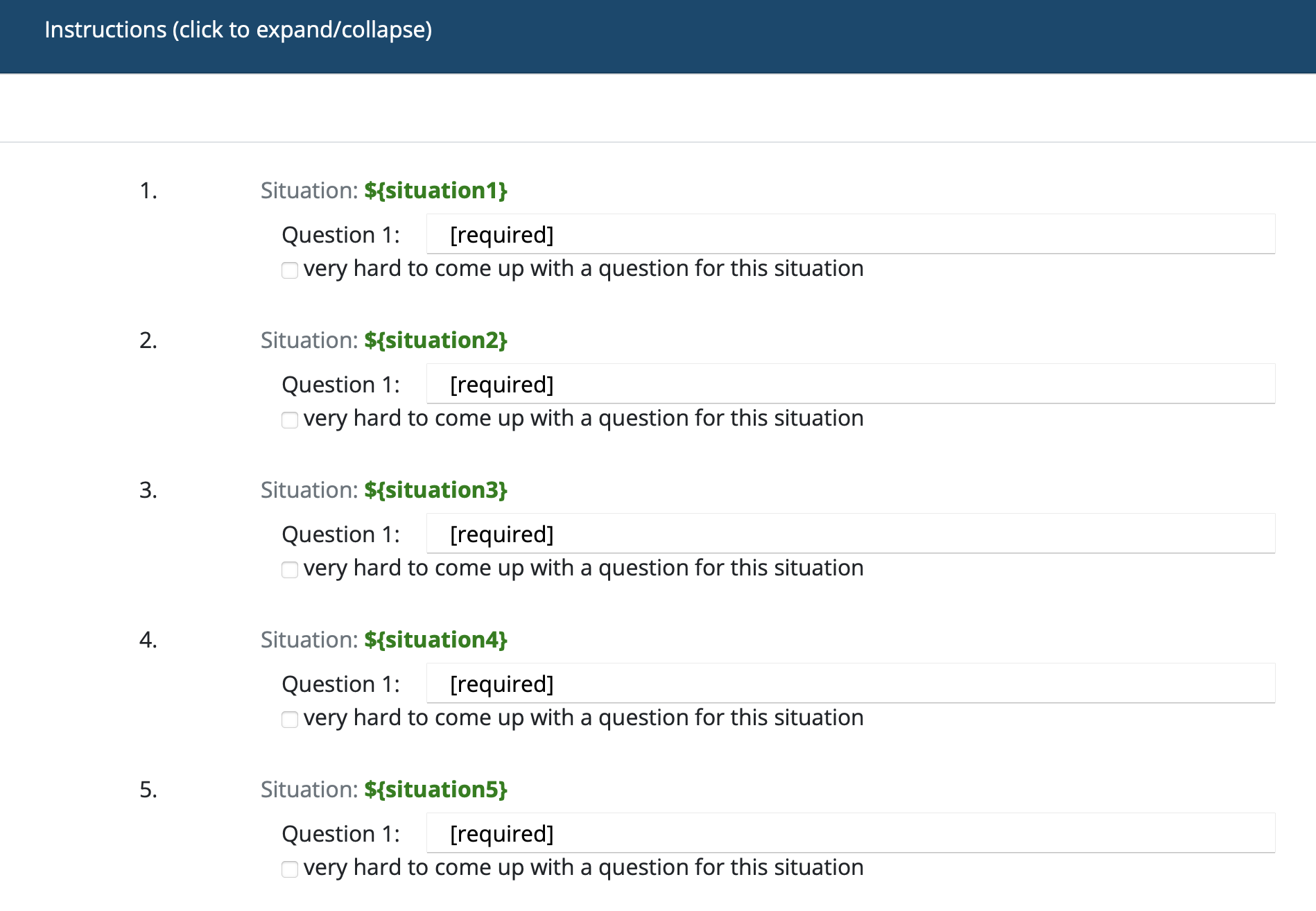}
    \caption{\dataset question collection interface.}
    \label{fig:hit-question-collection}
\end{figure*}

\paragraph{Question Collection:} We crowdsource clarification question by prompting annotators with a situation. The crowdworkers are asked to imagine a hypothetical situation where a colleague came to them requesting advice or judgment on the shown situation. The workers are then instructed to write a clarification question they would want to ask that would help them make a better judgment or give a better advice that they would without it. Each of the 6425 situation is presented to 5 distinct annotators; we collect 5 questions per situation. A screenshot of the HIT is shown in Figure~\ref{fig:hit-question-collection}.

\paragraph{Human Evaluation:} We ask crowdworkers to evaluate model outputs. Given a situation and a question Turkers are asked to rate the question along three different attributes:

\begin{itemize}[label={},noitemsep]
\item \textbf{Grammaticality} Is the question grammatical? - \textit{yes/no}
\item \textbf{Relevance} Does the question fit the situation and is it plausible that someone might ask this question? - \textit{very relevant/somewhat relevant/entirely irrelevant}
\item \textbf{Informativeness} Can the question lead to new information or does it ask about things already mentioned in the situation? - \textit{very/somewhat/uninformative}
\end{itemize}

\begin{figure*}[h!]
    \centering
    \includegraphics[width=0.9\textwidth]{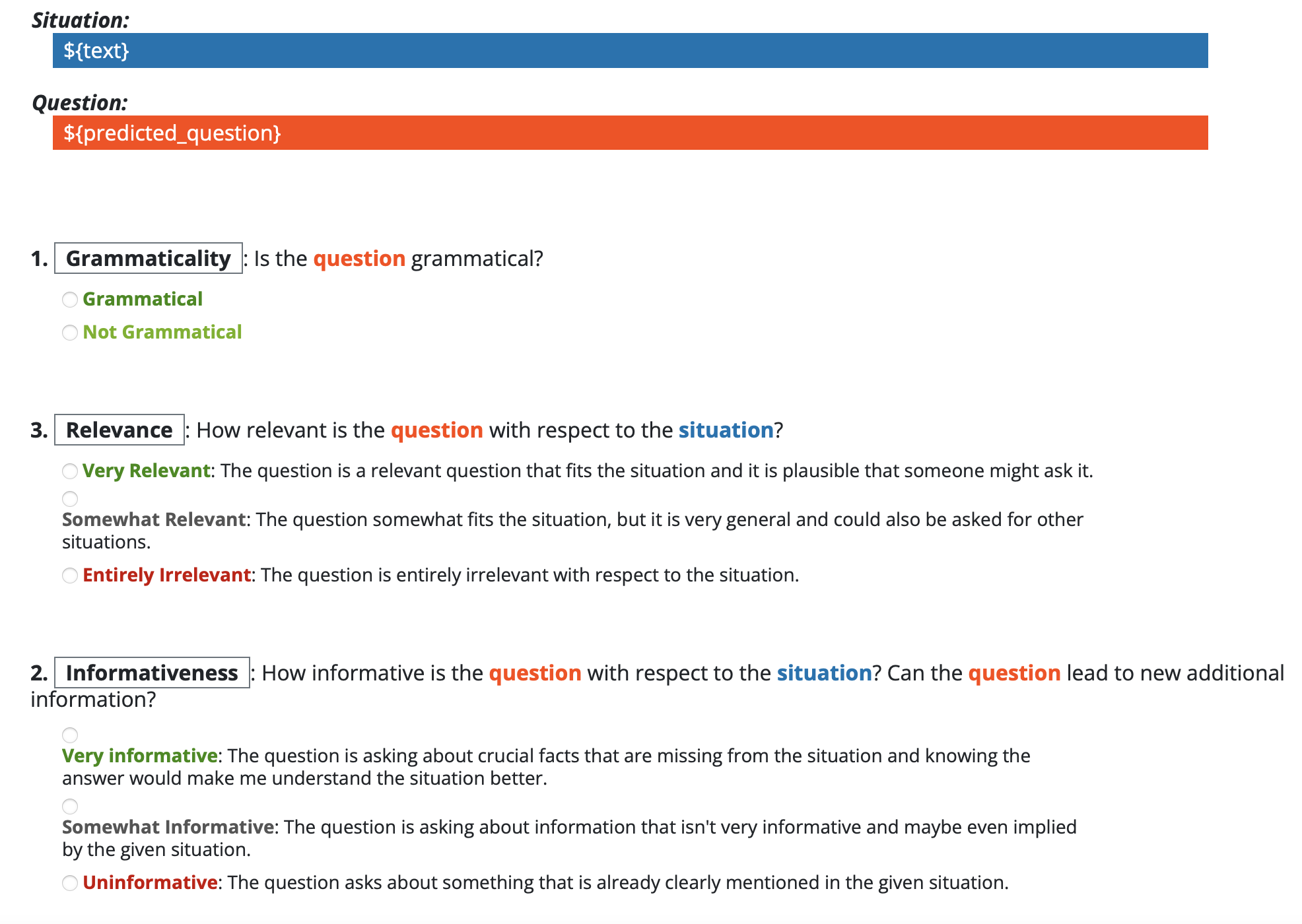}
    \caption{Informativeness, relevance and grammaticality evaluation interface.}
    \label{fig:hit-validity}
\end{figure*}

\begin{figure*}[h!]
    \centering
    \includegraphics[width=0.9\textwidth]{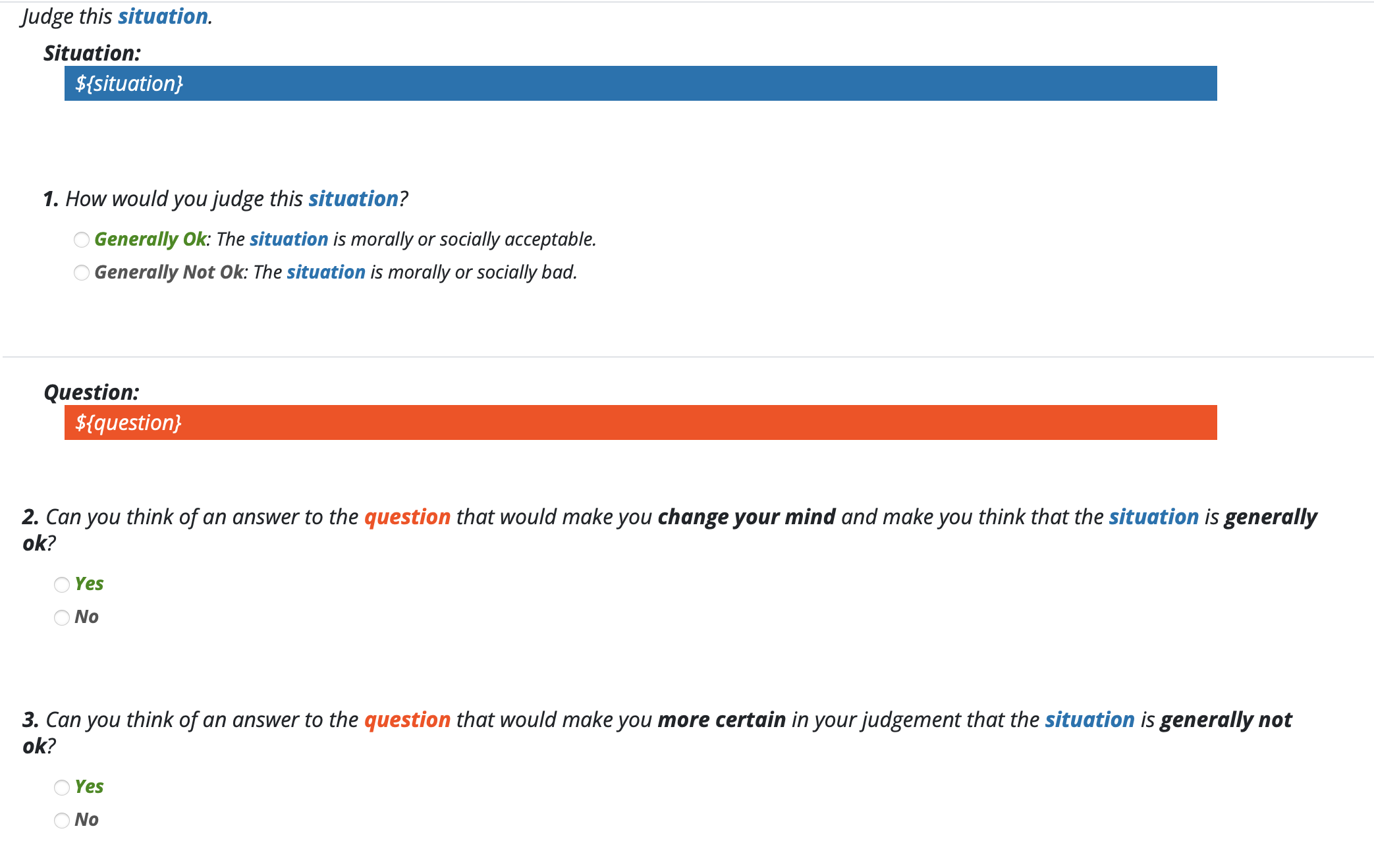}
    \caption{Defeasibility evaluation interface.}    \label{fig:hit-defeasibility}
\end{figure*}

A screenshot for the evaluation HIT w.r.t grammaticality, relevance, and informativeness is found in Figures~\ref{fig:hit-validity}. Additionally, we evaluate the \textbf{defeasibility} of a question. A screenshot of the defeasibility evaluation is shown in Figure~\ref{fig:hit-defeasibility}.

\paragraph{IRB approval:} We sought and received exemption from our internal IRB. In accordance to the regulations, we do not collect sensitive information. If we do publish WorkerIDs, we will do so by fully anonymizing the information. The exemption received does not require a consent form. 

\paragraph{Language and Demographics:} We have not collected any demographic information from the workers. However, all crowdsourcing was conducted in English and the region (current location of the crowdworker) was set to US. Consequently, what counts as a context of consequence is centered around western views, or views of the English speaking cultures within the United States.

\subsection{Prompting for Answer Generation}
One way to elicit a set of opposing answers is through prompting. We instruct GPT-3 to provide a so-called ``bad" and a so-called ``good" answer to a question about a situation. For the situation \sit{learning how to take a joke} and the question \ques{What was the joke?}, the two answers could be: \ans{it was a lighthearted joke among friends} and \ans{it was an offensive joke}. In order to determine which of the answers is a weakener and which a strengthener, we compare the difference in Delphi's judgement for $s$ and $s+a_{good}$ or $s+a_{bad}$.

\subsection{Details of PPO}
\paragraph{Policy loss.}
To compute the policy loss, we first define the \textit{truncated estimated advantage function},
\begin{align*}
& \hat{A}_t = \sum_{t'=t}^{T-1}{(\gamma \lambda)^{t'-t} \delta_{t'}}, \\
& \text{where} \quad \delta_{t'} = r_{t'} + \gamma V_\phi(s_{t'+1}) - V_\phi(s_{t'}),
\end{align*}
where the value function of a certain state $s_t$ is estimated by the value model $V_\phi(\cdot)$, $r_t$ denotes intermediate reward obtained at time step $t$, and $\gamma$ and $\lambda$ denote reward decay factor.
PPO then maximizes the empirical expectation of \textit{clipped surrogate objective} term, defined as
\begin{align*}
&\text{cso}(\hat{A}_t, r_t(\theta), \epsilon) = \\ &\qquad \min \big(r_t(\theta) \hat{A}_t, \text{clip}(r_t(\theta), 1 - \epsilon, 1 + \epsilon) \hat{A}_t \big),
\end{align*}

\paragraph{Value loss.}
The value model is trained with MSE loss with respect to the target value, $V^{\text{targ}}_t$:
\begin{align*}
& \mathcal{L}_{\text{Value}}(\phi) = \hat{E} \Big[ \big( V(s_t; \phi) - V^{\text{targ}}_t \big)^2 \Big], 
\end{align*}

\subsection{Implementation Details}
The question generation model $\theta_Q$ is trained with T5-large \cite{raffel2020exploring} on 
 \dataset. 

We train the answer generation model $\theta_A$ (T5-large) on 77k instances from $\dataset_{silver}$ (\S\ref{silver}) and about 4k instances obtained through prompting GPT-3. For the answer-filtering using NLI we use WaNLI \cite{liu-etal-2022-wanli} as an off-the-shelf NLI model.

We run the PPO algorithm for 6000 steps, with a batch size of 64, top\_p$=0.6$ and a temperature of 0.7 for the sampling policy. Training for 6000 step took approximately 2 days on 4 A6000, 48G, GPUs.

Our models are based on T5-large which has 770 million parameters. 

For the \textsc{Delphi} model we use the $t5-large$ variant.

\begin{figure*}[h!]
    \centering
    \includegraphics[width=0.9\textwidth]{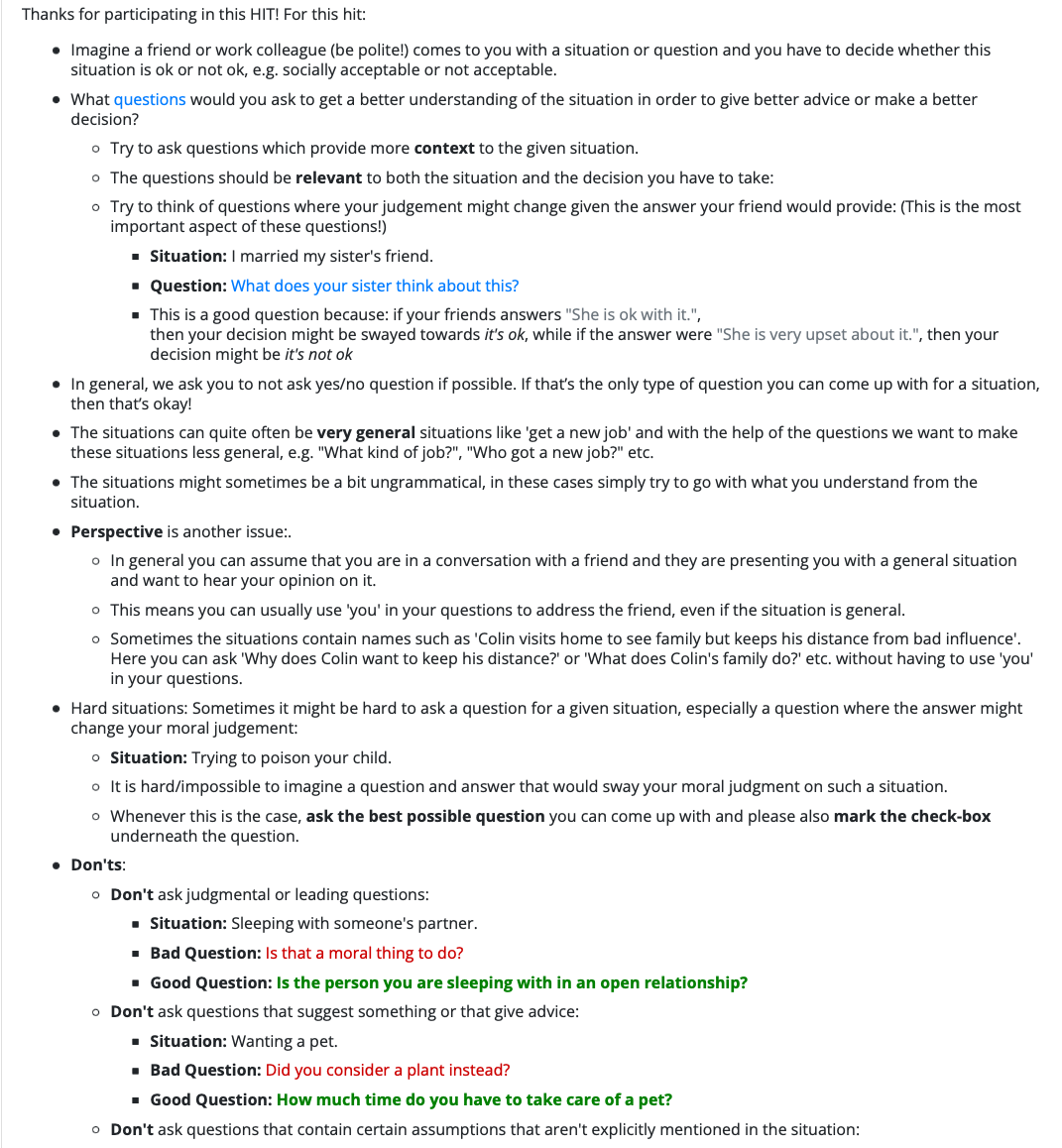}
    \caption{\dataset question collection instructions.}
    \label{fig:hit-question-collection-inst}
\end{figure*}

\begin{figure*}[h!]
    \centering
    \includegraphics[width=0.9\textwidth]{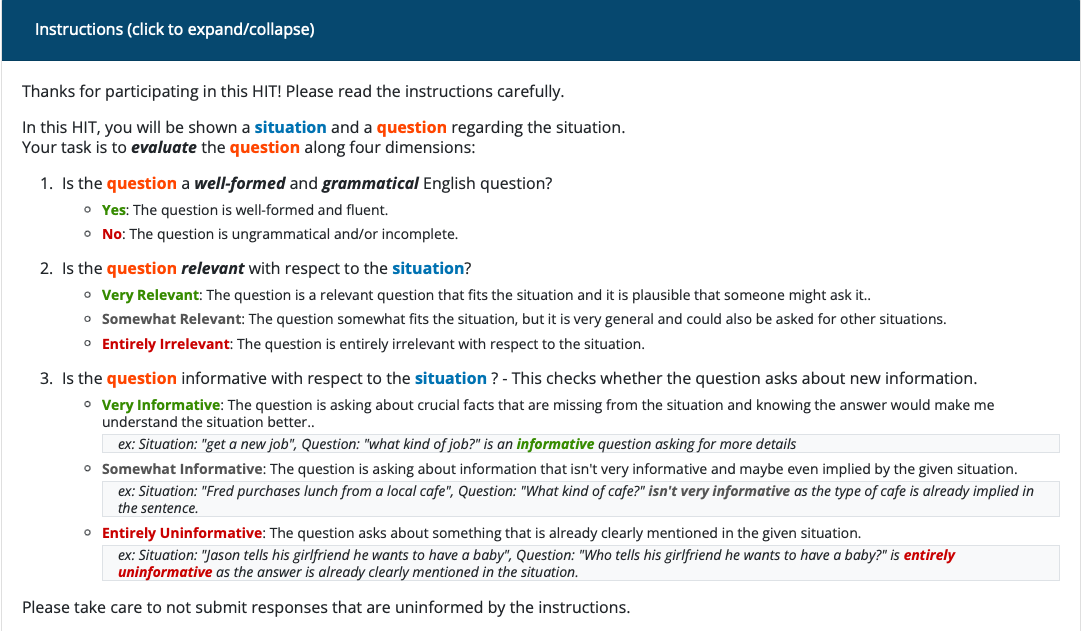}
    \caption{Informativeness, relevance and grammaticality evaluation instructions.}
    \label{fig:hit-validity-inst}
\end{figure*}

\begin{figure*}[h!]
    \centering
    \includegraphics[width=0.9\textwidth]{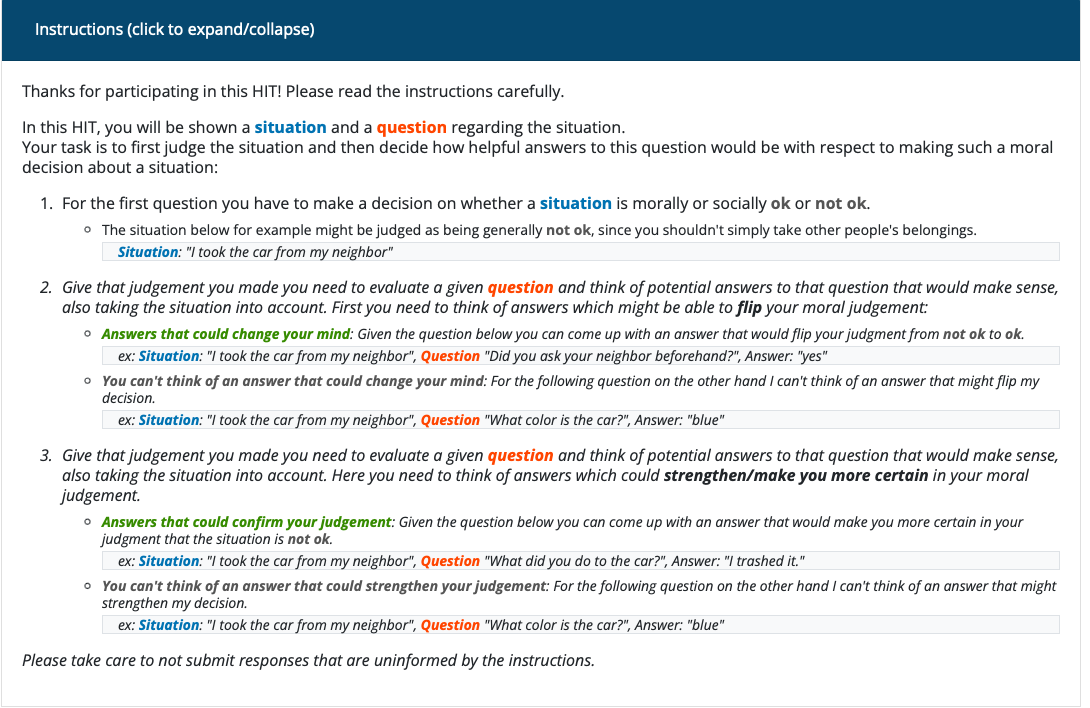}
    \label{fig:hit-defeasibility-instructions}
    \caption{Defeasibility evaluation instructions.}
\end{figure*}

\subsection{Licensing}
\model and the code will be released under the Apache License 2.0. We will release the dataset under CC-BY.

\subsection{Sample of Generated Questions}
We show a sample of 15 generated questions in Tab.~\ref{tab:gen}.
\begin{table*}[h!]
\begin{tabular}{l|l}
Situation                                                    & Question                                     \\ \hline
sleeping in when you have to get up.                         & What are you doing when you have to get up?  \\
Paying more when you are able to because you make more.      & What are you making?                         \\
Apologizing when you have done something wrong.              & What did you do?                             \\
being careful when opening your car door in a parking lot.   & Are you aware of what is inside your car?    \\
being out after curfew.                                      & Are you allowed to be out after curfew?      \\
respecting your friend's wishes                              & What is your friend's wish?                  \\
making a scene                                               & What kind of scene?                          \\
being friends with benefits with someone                     & Why are you friends with benefits with them? \\
not liking certain dogs                                      & Why don't you like them?                     \\
smoking in one's own home                                    & What are the risks?                          \\
wanting to go home with someone                              & How do you feel about them?                  \\
being angry with your partner when they hide things & What are they hiding?                        \\
Talking about things you know nothing about                  & Why are you talking about these things?      \\
agreeing to go out with someone who bullied you  & Why are they bullying you?                   \\
using your money how you see fit                             & What are you doing with your money?         
\end{tabular}
\caption{Sample of situations from the dev set of SocialChem and the \model generated question.}
\label{tab:gen}
\end{table*}

\subsection{Sample of Generated Questions in a Compositional Setting}
We show a sample of generated questions in a two-turn interactive setting in Tab.~\ref{tab:gen_inter}.
\begin{table*}[h!]
\begin{tabular}{l|l}
Situation                                                                                                                          & Question                                \\ \hline

daydreaming sometimes                                                                                                              & What are you daydreaming about?         \\
daydreaming sometimes about a vacation                                                                                             & What kind of vacation are you planning? \\
stealing money from your parents                                                                                                   & How did you steal the money?            \\
\begin{tabular}[c]{@{}l@{}}stealing money from your parents \\ by stealing it from their wallet\end{tabular}                       & How much money did you steal?           \\
getting some help when you're lost in life                                                                                         & Are you looking for help?               \\
\begin{tabular}[c]{@{}l@{}}getting some help when you're lost in life and\\ stating that you are not looking for help\end{tabular} & Are you having difficulty coping?       \\
loving your friend platonically                                                                                                    & What is your friend like?               \\
\begin{tabular}[c]{@{}l@{}}loving your friend platonically\\ because he is very nice\end{tabular}                                  & What do you like about your friend?     \\
being judgmental                                                                                                                   & How are you being judgmental?           \\
\begin{tabular}[c]{@{}l@{}}being judgmental because \\ I am silently judging them\end{tabular}                                     & What is the other person doing?         \\
getting cold feet before a wedding                                                                                                 & Why are you getting cold feet?          \\
\begin{tabular}[c]{@{}l@{}}getting cold feet before a wedding\\ because you only know your partner for 1 month\end{tabular}        & What is your relationship status?       \\
Being mindful of your health                                                                                                       & How are you being mindful?              \\
being mindful of your health by working out                                                                                        & What are the benefits of working out?  
\end{tabular}
\caption{Sample of situations from the dev set of SocialChem, their expansion through interaction (2 turns) and the \model generated question.}
\label{tab:gen_inter}
\end{table*}
\end{document}